\documentclass[10pt,twocolumn,letterpaper]{article}

\usepackage{cvpr}              

\usepackage[dvipsnames]{xcolor}

\definecolor{cvprblue}{rgb}{0.21,0.49,0.74}
\usepackage[pagebackref,breaklinks,colorlinks,citecolor=cvprblue]{hyperref}
\usepackage{multirow}
\usepackage{xcolor,colortbl}
\usepackage[normalem]{ulem}
\usepackage{pifont}
\newcommand{\cmark}{\ding{51}}%
\newcommand{\xmark}{\ding{55}}%
\usepackage{balance}
\usepackage{amsmath}
\usepackage{balance}
\usepackage{tabularx}
\def\paperID{10408} 
\def\confName{CVPR}
\def\confYear{2024}

\title{VidLA: Video-Language Alignment at Scale}

\author{Mamshad Nayeem Rizve ~~~~ Fan Fei ~~~~ Jayakrishnan Unnikrishnan ~~~~ Son Tran ~~~~ Benjamin Z. Yao\\Belinda Zeng ~~~~ Mubarak Shah ~~~~ Trishul Chilimbi\\
{\tt\small \{mnrizve, feiff, jayunn, sontran, benjamy, zengb, shahmub, trishulc\}@amazon.com}}

\newcommand{\cls}{\textsc{[cls]}\xspace}
\newcommand{\concept}{\textsc{[mst]}\xspace}
\newcommand{\patch}{\textsc{[patch]}\xspace}

\begin{document}
\maketitle
\begin{abstract}
In this paper, we propose VidLA, an approach for video-language alignment at scale. There are two major limitations of previous video-language alignment approaches. First, they do not capture both short-range and long-range temporal dependencies and typically employ complex hierarchical deep network architectures that are hard to integrate with existing pretrained image-text foundation models. To effectively address this limitation, we instead keep the network architecture simple and use a set of data tokens that operate at different temporal resolutions in a hierarchical manner, accounting for the temporally hierarchical nature of videos. By employing a simple two-tower architecture, we are able to initialize our video-language model with pretrained image-text foundation models, thereby boosting the final performance. Second, existing video-language alignment works struggle due to the lack of semantically aligned large-scale training data. To overcome it, we leverage recent LLMs to curate the largest video-language dataset to date with better visual grounding. Furthermore, unlike existing video-text datasets which only contain short clips, our dataset is enriched with video clips of varying durations to aid our temporally hierarchical data tokens in extracting better representations at varying temporal scales. Overall, empirical results show that our proposed approach surpasses state-of-the-art methods on multiple retrieval benchmarks, especially on longer videos, and performs competitively on classification benchmarks.
\end{abstract}

\section{Introduction}
\label{sec:intro}

\begin{figure}[ht!]
\begin{center}
\includegraphics[width=0.40\textwidth]{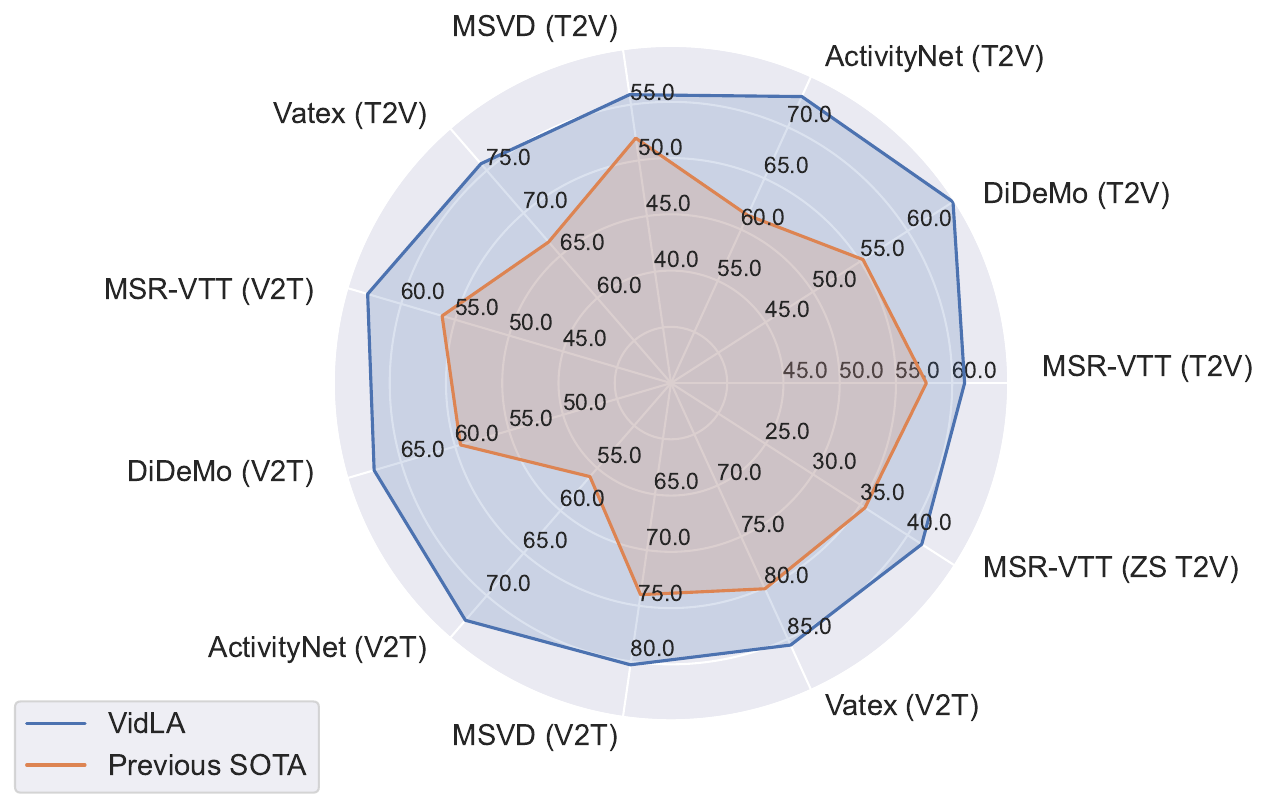}
\vspace{-1.0em}
\caption{Recall@1 performance on retrieval benchmarks compared to previous SoTA with ViT-B scale models.} 
\label{fig:radar}
\end{center}
\vspace{-8mm}
\end{figure}

Vision-language alignment is crucial for solving many vision-language tasks like text-guided retrieval~\cite{singh2022flava, radford2021learning, li2023blip, li2020unicoder, li2020oscar}, visual question answering~\cite{chen2022pali, li2022mplug, yu2022coca, wang2022image}, visual captioning~\cite{li2023blip, li2022mplug, wang2022ofa, nguyen2022grit}, etc. In the past few years, training on web-scale image-text pairs has significantly improved performance in many of these vision-language tasks~\cite{radford2021learning, li2021align, yu2022coca, li2022blip}. However, humans perceive the world as a continuous stream of images, which is also reflected in the increasing number of videos uploaded to the web daily. Despite the ubiquitous use of videos, when compared to image-language alignment, video-language alignment is significantly more challenging for two key reasons.

\textit{First}, unlike image-language data, it is much harder to collect aligned video-language data at scale. To address this issue, most prior works utilize automatic speech recognition (ASR) systems~\cite{radford2023robust, zhang2020pushing, amodei2016deep} to extract textual transcripts and generate paired video-language data for large-scale training~\cite{miech2019howto100m, xue2022advancing, zellers2022merlot}. However, it has been shown that transcripts often corresponds poorly with their associated visual contents ~\cite{shvetsova2023howtocaption, han2022temporal, miech2020end, miech2019howto100m}. As a result, some recent works~\cite{li2023unmasked, cheng2023vindlu, wang2022omnivl, lei2022revealing} skipped large-scale video-language training and worked around by utilizing language-aligned image encoders, followed by lightly adapting them with temporal modules  on small-scale video datasets with paired textual descriptions~\cite{bain2021frozen, nagrani2022learning}. However, training on such small-scale datasets often leads to overfitting~\cite{xue2022clip} and does not encourage learning temporally diverse representations~\cite{lei2022revealing}. 
\textit{Second}, since the vision transformer architecture lacks strong visual inductive bias such as that in CNN-type architectures, it requires large-scale pretraining data for effective generalization~\cite{dosovitskiy2020image, raghu2021vision}. In case of videos, this problem is amplified further due to the added temporal dimension. Facing this challenge, to be more efficient, existing works utilize factorized~\cite{arnab2021vivit, bertasius2021space, neimark2021video} or hierarchical~\cite{li2022mvitv2, liu2022video, li2022uniformer, ryali2023hiera} space-time attention. However, neither of these solutions are optimal for large-scale video-language alignment, as factorized space-time attention overly focuses on aggregating redundant local spatio-temporal information~\cite{li2022uniformer}, while hierarchical space-time attention makes the use of pretrained language-aligned non-hierarchical image encoders~\cite{radford2021learning, li2021align} challenging. Our work addresses both challenges in large-scale video-language alignment using large language models and a novel hierarchical temporal attention mechanism. 

\begin{figure}[ht!]
\begin{center}
\vspace{-2mm}
\includegraphics[width=0.9\columnwidth]{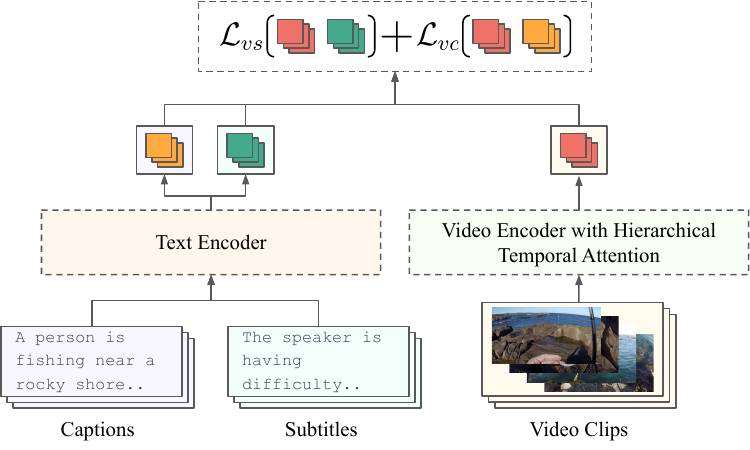}
\vspace{-4mm}
\caption{Figure summarizing our video-language alignment training approach with a two-tower architecture, where text encoder and video encoder with hierarchical temporal attention are trained with info-NCE losses to align video representations with subtitle and caption text representations simultaneously. We generate the captions using a multi-modal LLM and utilize an LLM to summarize the caption and subtitle texts.}
\label{fig:main}
\end{center}
\vspace{-1.5em}
\end{figure}

In addition, having large-scale video-text data set is crucial for video-language alignment training. Towards that end, we construct a very large dataset, with $\sim$800M video-text pairs, to train video-language alignment model at scale.    
In this context, we propose several simple data curation strategies using LLMs~\cite{touvron2023llama, chung2022scaling, chowdhery2022palm, chiang2023vicuna, brown2020language} to improve the semantic correlation between textual description and associated visual content of large-scale video-language datasets. First, we utilize recent  multi-modal large language models (MLLM)~\cite{li2023blip, liu2023visual, dai2023instructblip, zhu2023minigpt, ye2023mplug} to generate auxiliary captions grounded in visual content. Second, to scale our solution, we generate captions at a low frame rate, capitalizing on temporal redundancy of videos. Third, we augment the existing video-language datasets by incorporating videos of varying lengths to facilitate robust alignment across diverse temporal scales. We utilize LLMs for summarizing longer video descriptions, preventing training asymmetry when we sample the same number of frames for videos of all durations but use longer texts for longer videos. The LLM summarization also helps when long textual descriptions cover disparate concepts.    

To efficiently utilize the non-hierarchical image-text pretrained models while accounting for the temporally hierarchical nature of videos, we factorize the space-time attention operation into two parts: local and global temporal attention. First, we focus on modeling the \textit{local} temporal relationships by treating videos as collections of temporal tubes of single patch tokens. This attention operation focuses on capturing fine-grained motion across frames. Next, to model \textit{global} spatio-temporal relationships in a temporally \textit{hierarchical} manner, inspired from prior art~\cite{feichtenhofer2019slowfast, zhou2018temporal}, we incorporate several Multi-Scale Temporal  \concept tokens that interact with all patch tokens at varying temporal resolutions to summarize the video semantic concepts at various temporal scales. To make this space-time attention operation more efficient, we update the patch tokens by attending over all the \concept tokens and other patch tokens from the same frame. Finally, we utilize a \cls token to attentively aggregate information from all the \concept and patch tokens. We utilize this aggregated spatio-temporal representation for video-language alignment training.  Our hierarchical temporal attention design not only models local temporal relationships but also models global temporal relationships at different temporal hierarchies while utilizing strong pretrained image-text encoders.
 
To summarize, in this work, we make two major technical contributions: (i) we propose several techniques to utilize LLMs to generate a large scale video-text dataset where the generated text has high semantic correlation with the visual content. (ii) we design a hierarchical temporal modeling approach that effectively incorporates pretrained image-text encoders and handles videos of varying lengths in the training set to improve downstream performance as shown in Figure~\ref{fig:radar}. We extensively evaluate the performance of our method on several video-text retrieval benchmarks to demonstrate its efficacy and the effectiveness of our data curation techniques and modeling approach. A summary of our approach is provided in Figure~\ref{fig:main}.
\section{Related Works}
\begin{table*}[ht!]
    \centering
    \resizebox{0.9\textwidth}{!}{
    \begin{tabular}{lcc|ccc|ccc}
\hline

\hline

\hline\\[-3mm]
        \multirow{2}{*}{Clip Duration} & \multirow{2}{*}{\# clips} & \multirow{2}{*}{Length (s)} &  \multicolumn{3}{c|}{Subtitle} & \multicolumn{3}{c}{Caption} \\
        & & & {\# sent} & {\# words} & {Summarized \# words} & {\# cap} & {\# words} & {Summarized \# words} \\
        \hline
        Short & 496M & 13.2 & 2.1 & 31.6 & 31.6* & 1.0 & 10.9 & 10.9*\\
        Medium & 212M & 30.4 & 4.7 & 72.1 & 19.5 & 2.3 & 24.8 & 15.2\\
        Long & 100M & 60.3 & 9.2 & 142.1 & 22.4 & 4.5 & 48.8 & 18.8 \\
\hline

\hline

\hline
\vspace{-1.7em}

    \end{tabular}}
    \caption{Statistics of our curated training data set YT-VidLA-800M. *For short video clips, texts are not summarized.}
    \label{tab:yt800m}
\vspace{-1.8em}
\end{table*}


\noindent{\bf Vision-Language Representation Learning} Recently, image-language models \cite{radford2021learning, jia2021scaling, yu2022coca, li2021align, yao2021filip, singh2022flava} have drawn huge attention because of their effectiveness in learning generic visual representations that are transferable to several downstream tasks like classification, retrieval, etc. This success can partly be attributed to the recently available large-scale image-text datasets \cite{schuhmann2021laion, schuhmann2022laion, 10.1145/2812802, desai2021redcaps, sharma2018conceptual}. However, in case of videos, there's no large-scale aligned video-language dataset. Therefore, to perform video-language pretraining, most recent works \cite{ni2022expanding, luo2022clip4clip, fang2021clip2video, gorti2022x, xue2022clip} bootstrap from a pretrained image-language model and then perform some form of lightweight adaptation on the video datasets. 

\noindent{\bf Adapting CLIP to Video} Many video foundation model works aim to adapt CLIP for video representation. Most use a straightforward approach and encode frame samples with CLIP image encoder and pool across frames with various mechanisms \cite{luo2022clip4clip, fang2021clip2video, gorti2022x} to represent the video. Other works insert additional temporal specific modelings such as divided space-time attention \cite{bertasius2021space} or adapters \cite{houlsby2019parameter} into CLIP vision transformer (ViT) layers \cite{cheng2023vindlu, wang2022omnivl, yang2023aim, pan2022st, ni2022expanding}. Among others there are also works using novel parallel architectures \cite{liu2023revisiting} or using addition special tokens to capture temporal interaction between frames \cite{xue2022clip}.

\noindent{\bf Video-Language Datasets}
For image-language pretraining, web images paired with alt-text have demonstrated to be extremely effective and can scale up to billions of samples \cite{schuhmann2022laion, wang2022git, chen2022pali, sun2023eva}. Video dataset using similar alt-text such as WebVid \cite{bain2021frozen} are often at a much smaller scale. Alternatively, VideoCC \cite{nagrani2022learning} dataset is generated by finding visually similar video clips from existing image text data. Video subtitle datasets on the other hand are much more widely available and easy to scale, leading to wide adoption \cite{miech2019howto100m, xue2022advancing, zellers2021merlot, zellers2022merlot}, however these type of videos are often very short clips segmented by sentence boundaries, and the subtitles are usually noisy and have poor visual grounding. In this work, instead of generating a new dataset, we propose a way to effectively use existing large scale video dataset to improve video text alignment.

\section{Video-Language Pretraining Dataset}
\label{sec:dataset}
A key component of our Video-Language Alignment method summarized in Figure~\ref{fig:main} is a high quality large scale video-language dataset. In order to be able to effectively train the video-text models with hierarchical temporal attention, and to allow the model to learn to align videos with different temporal scales, we need a dataset with videos of different lengths, and corresponding text annotations with varying levels of temporal abstraction. We describe our novel data curation scheme in detail below.

\noindent\textbf{Source Videos} We utilize 20 million videos from the YT-Temporal-1B~\cite{zellers2022merlot} dataset for creating video-text dataset since its the largest collection of publicly available videos. These videos cover a diverse set of video concepts, and have been filtered to ensure better visual groundedness as well as to protect user privacy. Unlike prior works which utilize the \textit{video frame}-subtitle pairs from this dataset, we create a new dataset composed of \textit{video clips} paired with subtitles and generated captions which we call YT-VidLA-800M. Next, we describe our multi-scale video clip extraction process. 

\noindent\textbf{Video Clip Extraction} To extract video clips, first, we punctuate the ASR subtitles using a bert-based~\cite{devlin2018bert} punctuation model to split the full subtitle into sentences. Next, we split the videos at sentence boundaries to generate clips, where each clip covers one or more sentences. To facilitate video-language alignment across different temporal scales, we split each video into clips of varying temporal lengths. To be particular, our shortest clips are on average 13 seconds long in duration, which are similar in length (6-13 seconds) to videos in existing large-scale datasets~\cite{xue2022advancing, miech2019howto100m, nagrani2022learning}. The medium length clips on the other hand are on average 30 seconds long, which is similar in length to videos in common retrieval benchmarks~\cite{anne2017localizing, xu2016msr}. The longer clips are on average 60 seconds in duration. Overall, we extract around 500 million short video clips, 200 million medium length video clips, and 100 million long video clips as summarized in Table~\ref{tab:yt800m}. Next, we discuss about our visually grounded auxiliary caption generation process.

\noindent\textbf{Caption Generation} To improve visual grounding in language supervision, we generate auxiliary captions for each clip using multi-modal LLMs. Particularly, we utilize BLIP-2~\cite{li2023blip} to generate captions for the frames of the extracted video clips. To be efficient, capitalizing on the temporal redundancy of videos, we generate these captions at a very low frame-rate ($\sim0.1$ FPS). To aggregate the frame-level captions to generate the video clip-level caption, we perform text summarization, which we discuss next.

\noindent\textbf{Text Summarization} We use an LLM~\cite{touvron2023llama} to summarize the individual frame captions to obtain the caption for each video clip. Additionally, we summarize ASR subtitles from longer videos to obtain right abstraction for video-language alignment training. Furthermore, caption and subtitle summarization address another practical problem: it reduces the size of the input text, making it feasible to fit within the context length window of CLIP's pretrained text encoder. After this operation, each video clip is paired with two summarized texts corresponding to ASR subtitle and generated caption. We present the statistics of YT-VidLA-800M before and after summarization in Table~\ref{tab:yt800m}.
\section{Method}

\begin{figure*}[ht]
\begin{center}
\includegraphics[width=0.7\textwidth]{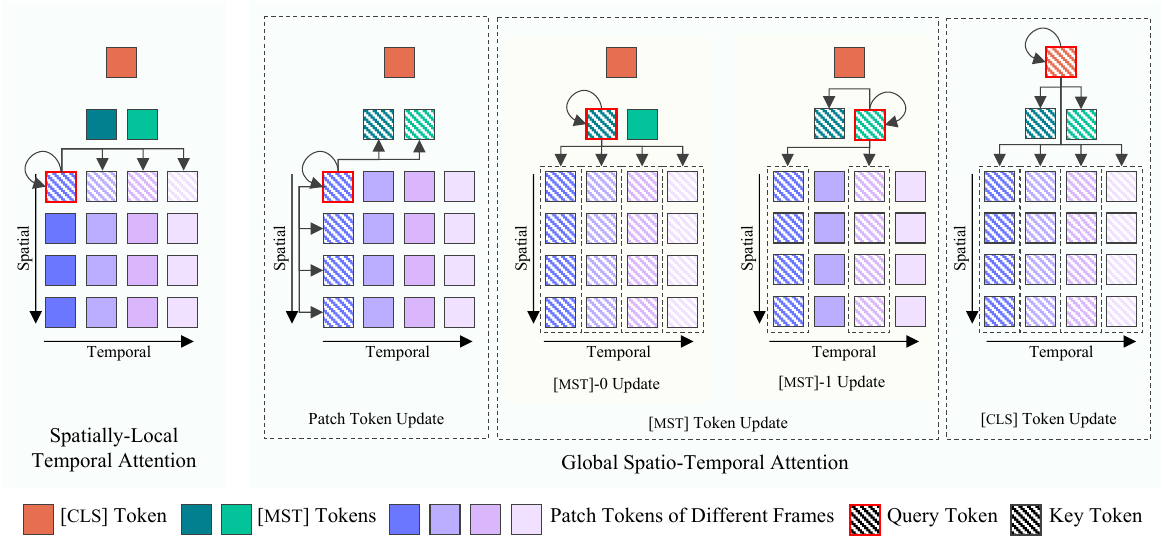}
\vspace{-2mm}
\caption{Figure summarizing the different tokens and the attention mechanisms used to update the tokens in our proposed Hierarchical Temporal Attention. This toy example uses $N=4$ patches, $T=4$ frames, $U=2$ levels of temporal hierarchy , $V=1$ \concept token per level and temporal scale $r=2$. Hierarchical temporal attention can be factorized into two parts. \textit{Spatially Local Temporal Attention (left):} Patch tokens only attend to its neighbors across time. For instance, \textit{first} patch token of the first frame gets updated by only attending to the \textit{first} patch token of all the other frames. \textit{Global Spatio-temporal Attention (right):} To capture global spatio-temporal semantics efficiently, we update the patch tokens by attending to other patch tokens from the same frame as well as all the \concept tokens. The \textit{third} and \textit{fourth} column depict the hierarchical \concept token update mechanism. Particularly, from the third column we observe that \concept-0 gets updated by attending to all the patch tokens and other \concept tokens of lower temporal resolution. The next column demonstrates the multi-scale \concept attention mechanism where the second \concept token, \concept-1, only attends to patch tokens from a subset of frames with a higher stride. The  \cls token acts as an \textit{aggregator} and attentively pulls information from both \concept and patch tokens.}
\label{fig:attention}
\end{center}
\vspace{-8mm}
\end{figure*}

In VidLA, we utilize an extension of the two-tower architecture for image-language alignment from CLIP~\cite{radford2021learning}. Particularly, we retain the CLIP text encoder architecture and extend CLIP's vision encoder to improve its temporal modeling capability by introducing a novel attention mechanism illustrated in Figure~\ref{fig:attention}. We provide details of our video encoder in the following.

\noindent\textbf{Preliminaries} The vision encoder accepts as input a video clip $\mathbf{v}$ consisting of $T$ RGB frames $\mathbf{v}^{t} \in \mathbb{R}^{H \times W \times 3}, t \in \{0, 1, ..., T-1\}$ each of size $H \times W$ pixels sampled from the video. Following vision transformer~\cite{dosovitskiy2020image} and pretrained image-language models~\cite{radford2021learning, li2021align}, we split each frame into non-overlapping patches of size $P\times P$ yielding $N=HW/P^2$ patches for each frame. Each of the $NT$ patches is linearly mapped with a learnable matrix and then combined with learnable spatial and temporal position embeddings to obtain a sequence of $TN$ patch tokens, represented by $\widetilde{\mathbf{Z}}^{0}\in\mathbb{R}^{TN\times d}$, where $d$ denotes the dimension of each token. We incorporate a set of $UV$ \concept tokens to capture summarized information at different temporal scales from the video where $U$ represents the number temporal hierarchies and $V$ represents the number of \concept tokens at each temporal hierarchy. We also include a \cls token  to capture the global representation of the video (see e.g., \cite{devlin2019bert}). We create the final input sequence, $\mathbf{Z}^{0}\in\mathbb{R}^{(1 + UV + TN)\times d}$, by prepending the learnable \cls token and $UV$ additional \concept tokens to the sequence of $TN$ patch tokens. The sequence of input tokens are passed through $L$ transformer layers. We use $\mathbf{Z}^{l}$ to denote the sequence after the $l$-th layer. In each layer the sequence is treated with two steps of attention followed by an MLP layer as summarized in Figure~\ref{fig:attention} and detailed next.

\noindent\textbf{Spatially Local Temporal Attention}
Inspired from a long line of works~\cite{simonyan2014two, feichtenhofer2019slowfast, feichtenhofer2016convolutional} that seek to model finegrained temporal motion for video understanding, we employ spatially local temporal attention. As the first operation in any $l$-th layer of the transformer, we 
remove the \cls and \concept tokens from the sequence of input tokens to that
layer, ${\mathbf{Z}}^{l-1}$, to apply this attention only on the patch tokens, $\widetilde{\mathbf{Z}}^{l-1}\in\mathbb{R}^{TN\times d}$. To capture finegrained temporal motion during this attention operation, each patch token only attends to patch tokens from other frames in the same spatial position, effectively allowing attention only along the temporal dimension. This operation can be represented using an attention mask, $\widetilde{\mathbf{M}} \in \mathbb{R}^{TN\times TN}$, formally defined as
\begin{align}
\widetilde{\mathbf{M}}_{i, j} = \notag
\begin{cases}
      0 & \text{if $\text{mod}(|j-i|,N) = 0$}\\
      -\infty & \text{otherwise.}
    \end{cases}
\end{align}
Spatially local temporal attention is then performed as
\begin{align}
\widetilde{\mathbf{Z}}^{l}_{SlT} = \mathrm{MMSA}(\mathrm{LN}(\widetilde{\mathbf{Z}}^{l-1}), \widetilde{\mathbf{M}}) + \widetilde{\mathbf{Z}}^{l-1}
\end{align}
where $\mathrm{LN}(.)$ is layer normalization~\cite{ba2016layer} operation and $\mathrm{MMSA}(.,.)$ is masked multi-head self-attention which can be expressed as $\mathrm{MMSA}(\mathbf{Z}, \mathbf{M}) := \mathrm{softmax}({\mathbf{Q}}{\mathbf{K}}^T/\sqrt{d} + \mathbf{M}) {\mathbf{V}} \in \mathbb{R}^{TN\times d}$; here $\mathbf{Q}, \mathbf{K}, \mathbf{V}$ are query, key, value embeddings of the sequence of input tokens $\mathbf{Z}$ obtained through linear projection and $\mathbf{M}$ is the input attention mask.

After the attention computation, we again prepend the \cls and \concept tokens to the updated patch tokens, $\widetilde{\mathbf{Z}}^{l}_{SlT}$, to obtain the 
token sequence ${\mathbf{Z}}^{l}_{SlT} = [(\mathbf{Z}^{l-1})^\cls, (\mathbf{Z}^{l-1})^\concept, \widetilde{\mathbf{Z}}^{l}_{SlT}]$.

\noindent\textbf{Global Spatio-Temporal Attention}
To efficiently model the global spatio-temporal semantics in a hierarchical manner, we utilize the hierarchical \concept tokens for guiding the global spatio-temporal attention. We employ an asymmetric attention mechanism to update the \cls, \concept, and patch tokens as illustrated in the second grid in Figure~\ref{fig:attention}. To keep the attention operation computationally efficient, each patch token attends to all patch tokens from the same frame, and to all the $UV$ \concept tokens $\in\mathbb{R}^{UV\times d}$. The patch token updates can be expressed using an attention mask, ${\mathbf{M}}^\patch \in \mathbb{R}^{TN\times (1 + UV + TN)}$, defined as 
${\mathbf{M}}^\patch = [\mathbf{0}, {\widetilde{ \mathbf{M}}}^G]$ where $\mathbf{0}$ is a $TN \times (1+UV)$ matrix of zeros and $\widetilde{\mathbf{M}}^G$ is a $TN \times TN$ matrix with
\begin{align}
{\widetilde{\mathbf{M}}}^G_{i,j} = \notag
\begin{cases}
      0 & \text{if $\left\lfloor\frac{i}{N}\right\rfloor = \left \lfloor\frac{j}{N}\right \rfloor$}\\      
      -\infty & \text{otherwise}
    \end{cases}
\end{align}
where $\lfloor . \rfloor$ indicates the FLOOR function. 

The update procedure for \concept tokens is designed to capture the temporally hierarchical nature of video concepts. The attention mask for each \concept token is determined by the hierarchy level of that token, ranging from $0$ to $U-1$, and the temporal scale $r$. Specifically, the \concept tokens from a particular hierarchy level $u$ attend to \concept tokens from lower temporal hierarchies and to the \patch tokens from every $r^u$-th frame. As there are $V$ \concept tokens in each hierarchy level, the updates for the \concept tokens can be expressed using another attention mask, ${\mathbf{M}}^\concept \in \mathbb{R}^{UV\times (1 + UV + TN)}$ where the first $V$ rows correspond to \concept tokens of hierarchy level $0$, followed by $V$ rows of hierarchy level $1$, and so on. The attention mask can be formally expressed as ${\mathbf{M}}^\concept = [-\infty\mathbf{1}, \widetilde{\mathbf{M}}^{{\concept,\mbox{self}}}, \widetilde{\mathbf{M}}^{{\concept,\mbox{patch}}}]$ where $\mathbf{1}$ is a $UV \times 1$ vector of all $1$'s, ${\mathbf{M}}^{{\concept,\mbox{self}}}$ is a $UV \times UV$ matrix and ${\mathbf{M}}^{{\concept,\mbox{patch}}}$ is a $UV \times TN$ matrix with
\begin{align}
{\widetilde{\mathbf{M}}}^{{\concept,\mbox{self}}}_{i,j} &= \notag
\begin{cases}
      0 & \text{if $\left\lfloor\frac{i}{V}\right\rfloor \geq \left \lfloor\frac{j}{V}\right \rfloor$}\\     
      -\infty & \text{otherwise}
    \end{cases}\\
{\widetilde{\mathbf{M}}}^{{\concept,\mbox{patch}}}_{i,j} &= \notag
\begin{cases}
      0 & \text{if $\mbox{mod}\left(\left\lfloor\frac{j}{N}\right\rfloor, r^{\left\lfloor\frac{i}{V}\right\rfloor}\right) = 0$}\\           
      -\infty & \text{otherwise}
    \end{cases}
\end{align}

Note that both patch and \concept tokens do not attend to the \cls token to limit propagation of \textit{global} information into the these \textit{local} tokens. We update the \cls token by attending to all the patch and \concept tokens.  This asymmetric update ensures that the \cls token merely acts as an \textit{aggregator} where it attentively pulls information from all tokens. We denote the attention mask for updating the \cls token as ${\mathbf{M}}^\cls \in \mathbb{R}^{1\times (1 + UV + TN)}$. We set all the entries of ${\mathbf{M}}^\cls$ to $0$ to allow attention computation with all tokens. Finally, we vertically stack these attention masks, $[{\mathbf{M}}^\cls, {\mathbf{M}}^\concept, {\mathbf{M}}^\patch]$, to generate the attention mask, $\mathbf{M}$, for global spatio-temporal attention. The global spatio-temporal attention mechanism also includes an MLP and skip connection as summarized in the following,
\begin{align}
\notag
{\mathbf{Z}}^{l}_{GST} &= \mathrm{MMSA}(\mathrm{LN}({\mathbf{Z}}^{l}_{SlT}), {\mathbf{M}})) + {\mathbf{Z}}^{l}_{SlT} \\
{\mathbf{Z}}^{l} &= \mathrm{MLP}({\mathbf{Z}}^{l}_{GST}) + {\mathbf{Z}}^{l}_{GST}
\end{align}
We propagate these updated token embeddings, ${\mathbf{Z}}^{l}$, to the next transformer layer. Finally, we use a linear projection of the \cls token from the last transformer layer as the video embedding for video-language alignment training.

\vspace{4mm}
\noindent\textbf{Pretraining Objective}
For video-language alignment training, we use language supervision from both ASR subtitle, $\mathbf{t}_s$, and caption, $\mathbf{t}_c$. Let's assume $s \in \mathbb{R}^D$, $c \in \mathbb{R}^D$ and $v \in \mathbb{R}^D$ are the encoded features vectors for subtitle, caption and video. We use the commonly used info-NCE loss~\cite{chen2020simple} as the objective function for video-language alignment training. The overall objective function is
\begin{align}
    \mathcal{L} = \frac{1}{B}\sum_{i = 1}^{B} (\mathcal{L}_{vs}^{i}) + \frac{1}{B}\sum_{i = 1}^{B} (\mathcal{L}_{vc}^{i})
\end{align}
\noindent where, $\mathcal{L}_{vs}$ and $\mathcal{L}_{vc}$ are info-NCE loss between video representation and the language representation from subtitle $s$ and caption $c$ respectively; for each loss,

\begin{align*}
\mathcal{L}_{vt}^{i}  \hspace{-0.02 in} = -\log\frac{\exp(v_i^\top t_i/\tau)}{\sum_{j=1}^{B}{\exp(v_i^\top t_j/\tau)}}  \hspace{-0.02 in} - \log\frac{\exp(t_i^\top v_i/\tau)}{\sum_{j=1}^{B}{\exp(t_i^\top v_j/\tau)}}
\end{align*}
\noindent where $t \in \{c, s\}$,  $B$ is the batch size and $\tau$ is the learnable temperature scale.
\begin{table*}[ht!]
\begin{center}\setlength{\tabcolsep}{10pt}
\resizebox{\textwidth}{!}{
\begin{tabular}{lllllll|llllll}
\hline

\hline

\hline\\[-3mm]
 \multicolumn{1}{l}{\multirow{2}{*}{{Method}}} &
 \multicolumn{6}{c|}{{MSR-VTT Text-to-Video}} & \multicolumn{6}{c}{{MSR-VTT Video-to-Text}} \\  
\multicolumn{1}{c}{} & {R@1} & {R@5} & {R@10} & {Avg} & {MdR}{{$\mathord{\downarrow}$}} & {MnR}{{$\mathord{\downarrow}$}} & {R@1} & {R@5} & {R@10} & {Avg} & {MdR}{{$\mathord{\downarrow}$}} & {MnR}{{$\mathord{\downarrow}$}}\\


\hline
ClipBERT~\cite{lei2021less}         & 22.0 & 46.8 & 59.9 & 42.9 & 6.0 & $-$   & $-$ & $-$ & $-$ & $-$ & $-$ & $-$ \\
Support Set~\cite{patrick2020support} & 30.1 & 58.5 & 69.3 & 52.6 & 3.0 & $-$   & 28.5 & 58.6 & 71.6 & 52.9 & 3.0 & $-$ \\
HD-VILA~\cite{xue2022advancing}     & 35.6 & 65.3 & 78.0 & 59.6 & 3.0 & $-$   & $-$ & $-$ & $-$ & $-$ & $-$ & $-$ \\
All-in-One~\cite{wang2023all}       & 37.9 & 68.1 & 77.1 & 61.0 & $-$ & $-$   & 37.5 & 66.1 & 77.4 & 60.3 & $-$ & $-$ \\
Frozen~\cite{bain2021frozen}        & 32.5 & 61.5 & 71.2 & 55.1 & 3.0 & $-$   & $-$ & $-$ & $-$ & $-$ & $-$ & $-$ \\
\hline
\rowcolor{gray!5} \multicolumn{13}{c}{CLIP-ViT-B/32} \\
CLIP4Clip~\cite{luo2022clip4clip}   & 44.5 & 71.4 & 81.6 & 65.8 & 2.0 & 15.3    & 42.7 & 70.9 & 80.6 & 64.7 & 2.0 & 11.6 \\
CenterCLIP~\cite{zhao2022centerclip}    & 44.2 & 71.6 & 82.1 & 66.0 & 2.0 & 15.1  & 42.8 & 71.7 & 82.2 & 65.6 & 2.0 & 10.9 \\
CLIP2TV~\cite{gao2021clip2tv}       & 46.1 & 72.5 & 82.9 & 67.2 & 2.0 & 15.2    & 43.9 & 73.0 & 82.8 & 66.6 & 2.0 & 11.1 \\
CAMoE*~\cite{cheng2021improving}    & 47.3 & 74.2 & 84.5 & 68.7 & 2.0 & 11.9    & 49.1 & 74.3 & 84.3 & 69.2 & 2.0 & 9.9 \\
DRL~\cite{wang2022disentangled}     & 47.4 & 74.6 & 83.8 & 68.6 & 2.0 & $-$    & 45.3 & 73.9 & 83.3 & 67.5 & 2.0 & $-$ \\
STAN*~\cite{liu2023revisiting}      & 49.0 & 74.8 & 83.5 & 69.1 & 2.0 & $-$    & $-$ & $-$ & $-$ & $-$ & $-$ & $-$ \\
PIDRo~\cite{Guan_2023_ICCV}        & 48.2 & 74.9 & 83.3 & 68.8 & 2.0 &  12.6    & 47.4 & 74.8 & 84.1 & 68.8 & 2.0 & 8.7 \\
Cap4Video~\cite{wu2023cap4video}    & 49.3 & 74.3 & 83.8 & 69.1 & 2.0 & 12.0    & 47.1 & 73.7 & 84.3 & 68.4 & 2.0 & 8.7 \\
UATVR*~\cite{Fang_2023_ICCV}         & 49.8 & 76.1 & 85.5 & 70.5 & 2.0 & 12.9    & 51.1 & 74.8 & 85.1 & 70.3 & 1.0 & 8.3 \\
CLIPViP~\cite{xue2022clip}          & 50.1 & 74.8 & 84.6 & 69.8 & 1.0 & $-$    & $-$ & $-$ & $-$ & $-$ & $-$ & $-$ \\
CLIPViP*~\cite{xue2022clip}         & 55.9 & 77.0 & 86.8 & 73.2 & 1.0 & $-$    & $-$ & $-$ & $-$ & $-$ & $-$ & $-$ \\
\rowcolor{blue!5} VidLA-B/32        & 55.6 & 79.7 & 86.9 & 74.1 & 1.0 & 11.4      & 55.1 & 79.9 & 88.0 & 74.3 & 1.0 & 6.9 \\
\rowcolor{blue!5} VidLA-B/32* & \textbf{60.9}{\tiny{\textcolor{teal}{$\mathord{\uparrow}5.0$}}}	& \textbf{81.6} & \textbf{89.4} & \textbf{77.3} & 1.0 & \textbf{8.7}      & \textbf{60.8}{\tiny{\textcolor{teal}{$\mathord{\uparrow}9.7$}}} & \textbf{82.4} & \textbf{89.1} & \textbf{77.4} & 1.0 & \textbf{6.3} \\
\hline
\rowcolor{gray!5} \multicolumn{13}{c}{CLIP-ViT-B/16} \\
BridgeFormer~\cite{ge2022bridging}  & 37.6 & 64.8 & 75.1 & 59.2 & 3.0 & $-$    & $-$ & $-$ & $-$ & $-$ & $-$ & $-$ \\
CLIP2TV~\cite{gao2021clip2tv}       & 49.3 & 74.7 & 83.6 & 69.2 & 2.0 & 13.5   & 46.9 & 75.0 & 85.1 & 69.0 & 2.0 & 10.0 \\
TS2-Net~\cite{liu2022ts2} & 49.4 & 75.6 & 85.3 & 70.1 & 2.0 & 13.5 & 46.6 & 75.9 & 84.9 & 69.1 & 2.0 & 8.9\\
Cap4Video~\cite{wu2023cap4video}    & 51.4 & 75.7 & 83.9 & 70.3 & 1.0 & 12.4   & 49.0 & 75.2 & 85.0 & 69.7 & 2.0 & 8.0 \\
DRL*~\cite{wang2022disentangled}    & 53.3 & 80.3 & 87.6 & 73.7 & 1.0 & $-$   & 56.2 & 79.9 & 87.4 & 74.5 & 1.0 & $-$ \\
STAN*~\cite{liu2023revisiting}      & 54.6 & 78.4 & 85.1 & 72.7 & 1.0 & $-$   & $-$ & $-$ & $-$ & $-$ & $-$ & $-$ \\
PIDRo*~\cite{Guan_2023_ICCV}        & 55.9 & 79.8 & 87.6 & 74.4 & 1.0 & 10.7   & 54.5 & 78.3 & 87.3 & 73.4 & 1.0 & 7.5 \\
UATVR*~\cite{Fang_2023_ICCV}         & 53.5 & 79.5 & 88.1 & 73.7 & 1.0 & 10.2  & 54.5 & 79.1 & 87.9 & 73.8 & 1.0 & 7.6 \\
CLIPViP~\cite{xue2022clip}          & 54.2 & 77.2 & 84.8 & 72.1 & 1.0 & $-$   & $-$ & $-$ & $-$ & $-$ & $-$ & $-$ \\
CLIPViP*~\cite{xue2022clip}         & 57.7 & 80.5 & 88.2 & 75.5 & 1.0 & $-$   & $-$ & $-$ & $-$ & $-$ & $-$ & $-$ \\
\rowcolor{blue!5} VidLA-B/16    & 58.0 & 81.1 & 87.8 & 75.6 & 1.0 & 10.4    & 56.1 & 80.5 & 88.7 & 75.1 & 1.0 & 6.8 \\
\rowcolor{blue!5} VidLA-B/16*   & \textbf{61.1}{\tiny{\textcolor{teal}{$\mathord{\uparrow}3.4$}}} & \textbf{83.8} & \textbf{90.4} & \textbf{78.4} & 1.0 & \textbf{8.1}     & \textbf{63.1}{\tiny{\textcolor{teal}{$\mathord{\uparrow}6.9$}}} & \textbf{84.7} & \textbf{90.8} & \textbf{79.5} & 1.0 & \textbf{6.1} \\
\hline
\rowcolor{gray!5} \multicolumn{13}{c}{Two Stage Models with Cross-Modal Fusion Re-Ranking} \\
\textsc{VindLU}{$\dagger$}\cite{cheng2023vindlu}   & 46.5 & 71.5 &  80.4 & 66.1 & $-$ & $-$  & $-$ & $-$ & $-$ & $-$ & $-$ & $-$ \\
UMT{$\dagger$}~\cite{li2023unmasked}  & 51.0 & 76.5 & 84.2 & 70.6 & $-$ & $-$     & 49.0 & 77.0 & 84.7 & 70.2 & $-$ & $-$ \\
InternVideo(ViT-L){$\dagger$}*~\cite{wang2022internvideo} & 55.2 & 79.6 & 87.5 & 74.1 & $-$ & $-$   & 57.9 & $-$ & $-$ & $-$ & $-$ & $-$ \\
\hline 

\hline

\hline
\end{tabular}
}
\end{center}
\vspace{-1.5em}
\caption{Retrieval performance on the MSR-VTT benchmark, metrics used are recall at (R@) 1, 5, 10, average recall (Avg), top candidate median rank (MdR) and mean rank (MnR). * indicates inference with dual-softmax. {$\dagger$} indicates two-stage method with candidate re-ranking. Performance delta is calculated against SoTA two-tower methods.}
\label{tab:compare_retrieval_msrvtt}
\vspace{-4mm}
\end{table*}

\begin{table*}[ht!]
\begin{center}
\resizebox{\textwidth}{!}{
\begin{tabular}{llll|lll|lll|lll}
\hline

\hline

\hline\\[-3mm]
 \multicolumn{1}{l}{\multirow{2}{*}{{Method}}} & \multicolumn{3}{c|}{{DiDeMo}} & \multicolumn{3}{c|}{{ActivityNet Captions}} & \multicolumn{3}{c|}{{MSVD}} & \multicolumn{3}{c}{{Vatex}}\\  
\multicolumn{1}{c}{} & {R@1} & {R@5} & {R@10} & {R@1} & {R@5} & {R@10} & {R@1} & {R@5} & {R@10} & {R@1} & {R@5} & {R@10} \\


\hline
ClipBERT~\cite{lei2021less}         & 20.4 & 48.0 & 60.8    & 21.3 & 49.0 & 63.5    & $-$ & $-$ & $-$       & $-$ & $-$ & $-$ \\
Support Set~\cite{patrick2020support} & $-$ & $-$ & $-$       & 29.2 & 61.6 & $-$     & 28.4 & 60.0 & 72.9    & 45.9 & 82.4 & 90.4 \\
HD-VILA~\cite{xue2022advancing}     & 28.8 & 57.4 & 69.1    & 28.5 & 57.4 & 94.0    & $-$ & $-$ & $-$       & $-$ & $-$ & $-$ \\
All-in-One~\cite{wang2023all}       & 32.7 & 61.4 & 73.5    & 22.4 & 53.7 & 67.7    & $-$ & $-$ & $-$       & $-$ & $-$ & $-$ \\
Frozen~\cite{bain2021frozen}        & 34.6 & 65.0 & 74.7    & $-$ & $-$ & $-$       & 33.7 & 64.7 & 76.3    & $-$ & $-$ & $-$ \\
\hline
\rowcolor{gray!5} \multicolumn{13}{c}{CLIP-ViT-B/32} \\
CLIP4Clip~\cite{luo2022clip4clip}   & 43.4 & 70.2 & 80.6    & 40.5 & 72.4 & $-$     & 46.2 & 76.1 & 84.6    & $-$ & $-$ & $-$ \\
CenterCLIP~\cite{zhao2022centerclip}  & $-$ & $-$ & $-$       & 43.9 & 74.6 & 85.8    & 47.6 & 77.5 & 86.0    & $-$ & $-$ & $-$ \\
CLIP2TV~\cite{gao2021clip2tv}       & 45.5 & 69.7 & 80.6    & 44.1 & 75.2 & $-$     & 47.0 & 76.5 & 85.1    & $-$ & $-$ & $-$ \\
CAMoE*~\cite{cheng2021improving}    & 43.8 & 71.4 & $-$     & 51.0 & 77.7 & $-$     & 49.8 & 79.2 & 87.0    & $-$ & $-$ & $-$ \\
DRL~\cite{wang2022disentangled}     & 47.9 & 73.8 & 82.7    & 44.2 & 74.5 & 86.1    & 48.3 & 79.1 & \textbf{87.3}    & 63.5 & \textbf{91.7} & \textbf{96.5} \\
STAN*~\cite{liu2023revisiting}      & 51.3 & 75.1 & 83.4    & $-$ & $-$ & $-$       & $-$ & $-$ & $-$       & $-$ & $-$ & $-$ \\
PIDRo*~\cite{Guan_2023_ICCV}         & 48.6 & 75.9 & 84.4    & 44.9 & 74.5 & 86.1    & 47.5 & 77.5 & 86.0    & $-$ & $-$ & $-$ \\
UATVR~\cite{Fang_2023_ICCV}         & 43.1 & 71.8 & 82.3    & $-$ & $-$ & $-$       & 46.0 & 76.3 & 85.1    & 61.3 & 91.0 & 95.6 \\
CLIPViP~\cite{xue2022clip}          & 48.6 & 77.1 & 84.4    & 51.1 & 78.4 & 88.3    & $-$ & $-$ & $-$       & $-$ & $-$ & $-$ \\
CLIPViP*~\cite{xue2022clip}         & 53.8 & 79.6 & 86.5    & 59.1 & 83.9 & 91.3    & $-$ & $-$ & $-$       & $-$ & $-$ & $-$ \\
\rowcolor{blue!5} VidLA-B/32        & 56.9 & 82.2 & 89.2      & 61.3 & 84.8 & 91.3  & 48.6 & 77.9 & 85.7   & 66.5 & 86.2 & 88.4 \\
\rowcolor{blue!5} VidLA-B/32*        & \textbf{62.2}{\tiny{\textcolor{teal}{$\mathord{\uparrow}8.4$}}} & \textbf{84.6} & \textbf{90.0}         & \textbf{69.2}{\tiny{\textcolor{teal}{$\mathord{\uparrow}10.1$}}} & \textbf{88.2} & \textbf{93.3} & \textbf{52.7}{\tiny{\textcolor{teal}{$\mathord{\uparrow}2.9$}}} & \textbf{80.4} & 87.0   & \textbf{73.7}{\tiny{\textcolor{teal}{$\mathord{\uparrow}7.2$}}} & 87.6 & 89.1 \\
\hline
\rowcolor{gray!5} \multicolumn{13}{c}{CLIP-ViT-B/16} \\
BridgeFormer~\cite{ge2022bridging}  & 37.0 & 62.2 & 73.9    & $-$ & $-$ & $-$       & 52.0 & \textbf{82.8} & \textbf{90.0}    & $-$ & $-$ & $-$ \\
DRL~\cite{wang2022disentangled} & 49.0 & 76.5 & 84.5    & 46.2 & 77.3 & 88.2    & 50.0 & 81.5 & 89.5    & 65.7 & 92.6 & 96.7\\
UATVR~\cite{Fang_2023_ICCV}         & 45.8 & 73.7 & 83.3     & $-$ & $-$ & $-$       & 49.7 & 79.0 & 87.3    & 64.5 & 92.6 & 96.8 \\
Cap4Video~\cite{wu2023cap4video}    & 52.0 & 79.4 & 87.5    & $-$ & $-$ & $-$       & 51.8 & 80.8 & 88.3    & 66.6 & \textbf{93.1} & \textbf{97.0} \\
CLIPViP~\cite{xue2022clip}      & 50.5 & 78.4 & 87.1    & 53.4 & 81.4 & 90.0    & $-$ & $-$ & $-$   & $-$ & $-$ & $-$ \\
CLIPViP*~\cite{xue2022clip} & 55.3 & 82.0 & 89.3    & 61.4 & 85.7 & 92.6    & $-$ & $-$ & $-$   & $-$ & $-$ & $-$ \\
\rowcolor{blue!5} VidLA-B/16    & 61.1 & 83.7 & 89.1    & 65.2 & 87.4 & 92.8 & 51.5 & 79.9 & 86.9     & 69.2 & 87.1 & 88.9 \\
\rowcolor{blue!5} VidLA-B/16*   & \textbf{64.8}{\tiny{\textcolor{teal}{$\mathord{\uparrow}6.9$}}} & \textbf{86.0} & \textbf{91.8}    & \textbf{73.0}{\tiny{\textcolor{teal}{$\mathord{\uparrow}10.8$}}} & \textbf{89.9} & \textbf{93.6} & \textbf{55.9}{\tiny{\textcolor{teal}{$\mathord{\uparrow}3.9$}}} & 82.3 & 88.3     & \textbf{75.8}{\tiny{\textcolor{teal}{$\mathord{\uparrow}9.2$}}} & 88.3 & 89.3 \\
\hline
\rowcolor{gray!5} \multicolumn{13}{c}{Two Stage Models with Cross-Modal Fusion Re-Ranking} \\
\textsc{VindLU}{$\dagger$}\cite{cheng2023vindlu}   & 61.2 & 85.8 & 91.0    & 55.0 & 81.4 & 89.7    & $-$ & $-$ & $-$       & $-$ & $-$ & $-$ \\
UMT{$\dagger$}~\cite{li2023unmasked}  & 61.6 & 86.8 & 91.5   & 58.3 & 83.9 & 91.5    & 71.9 & 94.5 & 97.8    & $-$ & $-$ & $-$\\
InternVideo(ViT-L)*~\cite{wang2022internvideo} & 57.9 & 82.4 & 88.9    & 62.2 & 85.9 & 93.2    & 58.4 & 84.5& 90.4 & 71.1 & $-$ & $-$\\
\hline 

\hline

\hline
\end{tabular}
}
\end{center}
\vspace{-1.5em}
\caption{Text-to-video Retrieval performance on the DiDeMo, ActivityNet Captions, MSVD and Vatex datasets. * indicates inference with dual-softmax. {$\dagger$} indicates two-stage method with candidate re-ranking. Performance delta is calculated against SoTA two-tower methods.}
\label{tab:compare_retrieval}
\vspace{-1.2em}
\end{table*}

\section{Experiments and Results}
\noindent\textbf{Implementation Details} We initialize our text and video encoders form pretrained OpenAI CLIP~\cite{radford2021learning} checkpoints. We randomly initialize the \concept tokens. To ensure that the initializion of our video encoder is close to CLIP's vision encoder, we initialize the projection matrices of spatially local temporal attention with zero. During training, we uniformly sample 12 frames from each video clip. We use multi-scale random crop \cite{wang2018temporal} with a ratio of $1.0$ and $0.8$ to crop the video to $224\times224$ while preserving aspect ratio. We also apply random horizontal flip for augmentation. We train our models for 3 epochs. We use a initial learning rate of $2e-5$ with cosine decay to $4e-8$. For training, we utilize 128 A100 GPUs and set the batch size to 4096. We set the number of hierarchies, $U$, to 3, the number of \concept tokens in each hierarchy, $V$, to 4, and the temporal scale $r$ to 2. We provide additional training and finetuning implementation details in the Supplementary.

\noindent\textbf{Video-Text Retrieval Datasets}
We evaluate our retrieval performance on MSR-VTT~\cite{xu2016msr}, DiDeMo~\cite{anne2017localizing}, ActivityNet Captions~\cite{krishna2017dense}, MSVD~\cite{chen2011collecting} and   VATEX~\cite{wang2019vatex} datasets. On all these datasets, we finetune on the standard training split and test it on the standard test/val splits. Following prior works~\cite{wu2023cap4video, lei2021less, xue2022clip, bain2021frozen}, we concatenate the multiple descriptions to form a paragraph and perform paragraph-to-video retrieval on DiDeMo and ActivityNet Captions datasets. 

\noindent\textbf{Main Results} We compare the retrieval performance of our proposed method VidLA with other recent works on MSR-VTT, DideMo, ActivityNet Captions, MSVD, and VATEX datasets and report the results in Table~\ref{tab:compare_retrieval_msrvtt} and \ref{tab:compare_retrieval}. We use VidLA-X/Y to denote the variant of our model that uses ViT-X/Y as the vision encoder, e.g., VidLA-B/32 uses ViT-B/32 as the vision encoder. We present results with and without using dual-softmax~\cite{cheng2021improving} for score normalization prior to ranking at the inference stage. Our proposed method outperforms all prior works using a similar ViT backbone by a significant margin. Particularly, from results reported in Table~\ref{tab:compare_retrieval_msrvtt}, we observe that VidLA-B/32 outperforms the second best method, CLIP-ViP, by 5.5\% on MSR-VTT for text-to-video retrieval in terms of R@1 without dual-softmax. We notice similar improvement (3.8\%) with ViT-B/16 backbone. We also notice a large improvement on the video-to-text retrieval task. Table~\ref{tab:compare_retrieval}, demonstrates a similar pattern on other four datasets. Particularly, we observe a larger improvement on datasets with longer videos such as ActivityNet Captions and DiDeMo, where our proposed method outperforms the second best method, CLIP-ViP, by 8.4\% and 10.1\% respectively. These results demonstrate that our proposed method not only outperforms the prior best method but also attains larger improvement if the downstream dataset is temporally longer.
\section{Analysis and Discussion}

We empirically validate our design choices on the model architecture, dataset temporal scales, language supervision as well as their combined effect by conducting a series of experiments to evaluate the model's retrieval performance. In all experiments, unless otherwise specified, we use the VidLA-B/32 model pretrained on an 80M subset of the YT-VidLA-800M dataset for 1 epoch, finetuned on MSR-VTT dataset. For these analysis experiments, we evaluate the retrieval performance without DSL. This 80M subset is constructed by sampling about 2M random source videos and then splitting them into short, medium and long clips as discussed in Section~\ref{sec:dataset}. For a fair comparison with other methods, we also utilize the same ViT-B/32 model as the vision encoder, initialized from the same CLIP checkpoint, and trained with the same compute and data budget.

\begin{table}[h]
\begin{center}
\small
\resizebox{0.48\textwidth}{!}{%
\begin{tabular}{ccc|cccc}
\hline

\hline

\hline\\[-3mm]
{\multirow{2}{*}{\concept}} & {\multirow{2}{*}{Hierarchy}} & {\multirow{2}{*}{Local}} & \multicolumn{4}{c}{MSR-VTT Retrieval}\\
 & & & {R@1} & {R@5} & {R@10} & Avg\\

\hline

\xmark & \xmark & \xmark & 49.1 & 75.3 & 83.5 & 69.3 \\
\cmark & \xmark & \xmark & 49.2 & 77.6 & 85.2 & 70.7 \\
\cmark & \cmark & \xmark & 50.0 & 77.6 & 85.4 & 71.0 \\
\cmark & \xmark & \cmark & 51.3 & 76.5 & 85.0 & 70.9 \\
\rowcolor{blue!5} \cmark & \cmark & \cmark & \textbf{53.5} & 77.5	& 85.6	& \textbf{72.2} \\
\hline 

\hline

\hline
\end{tabular}
}
\end{center}
\vspace{-1.2em}
\caption{Comparison of retrieval performances on MSR-VTT dataset with different settings for \concept token attention and the effect of spatially-local temporal attention.}
\label{tab:multi_scale_attn}
\vspace{-1.0em}
\end{table}


\begin{table}[h]
\begin{center}
\small
\begin{tabular}{c|cccc}
\hline

\hline

\hline\\[-3mm]
{\multirow{2}{*}{Multi-Scale}} & \multicolumn{4}{c}{MSR-VTT Retrieval}\\
 & {R@1} & {R@5} & {R@10} & Avg\\
\hline
\xmark & 51.9 & 78.2 & 85.6 & {71.9} \\
\rowcolor{blue!5} \cmark & \textbf{53.5} & 77.5	& 85.6	& \textbf{72.2}  \\
\hline 

\hline

\hline
\end{tabular}
\end{center}
\vspace{-1.2em}
\caption{Ablation study on the length distribution of videos in the pretraining dataset. Retrieval performance improves when the dataset is created with \textit{short}, \textit{medium} and \textit{long} clips}
\label{tab:multi_scale_data}
\vspace{-6mm}
\end{table}

\noindent\textbf{Attention Design} To analyze the effectiveness of \concept guided hierarchical temporal attention mechanism, we conduct a series of experiments with different attention configurations and report the results in Table~\ref{tab:multi_scale_attn}. The first two rows demonstrate the effectiveness of \concept tokens, even without any temporal hierarchy. Third row demonstrates the effectiveness of introducing multiple temporal hierarchies in \concept tokens. On the other hand, the fourth row shows the effectiveness of spatially-local temporal attention, where it provides a significant improvement in terms of R@1 retrieval performance over the seon. Finally, the last row confirms the efficacy of our proposed temporal attention mechanism, providing a substantial 4.4\% improvement over the baseline. Overall, these results not only validate the effectiveness of our proposed attention mechanism but also highlight the efficacy of its individual components.  

\noindent\textbf{Temporal Scales in Pretraining Data} To analyze the impact of incorporating multiple temporal scales in the proposed pretraining dataset, we compare a model pretrained on the 80M subset containing \textit{short}, \textit{medium} and \textit{long} clips against a model trained on only \textit{short} short clips from the same set of 2M videos. For a fair comparison, we train these models for same number of steps. We present the finetuned results in Table~\ref{tab:multi_scale_data} and observe that including multiple scales in the pretraining dataset helps boost retrieval performance.

\begin{figure}[t]
\captionsetup[subfloat]{labelformat=empty}
    \centering
    \subfloat[]{{\includegraphics[width=0.24\textwidth]{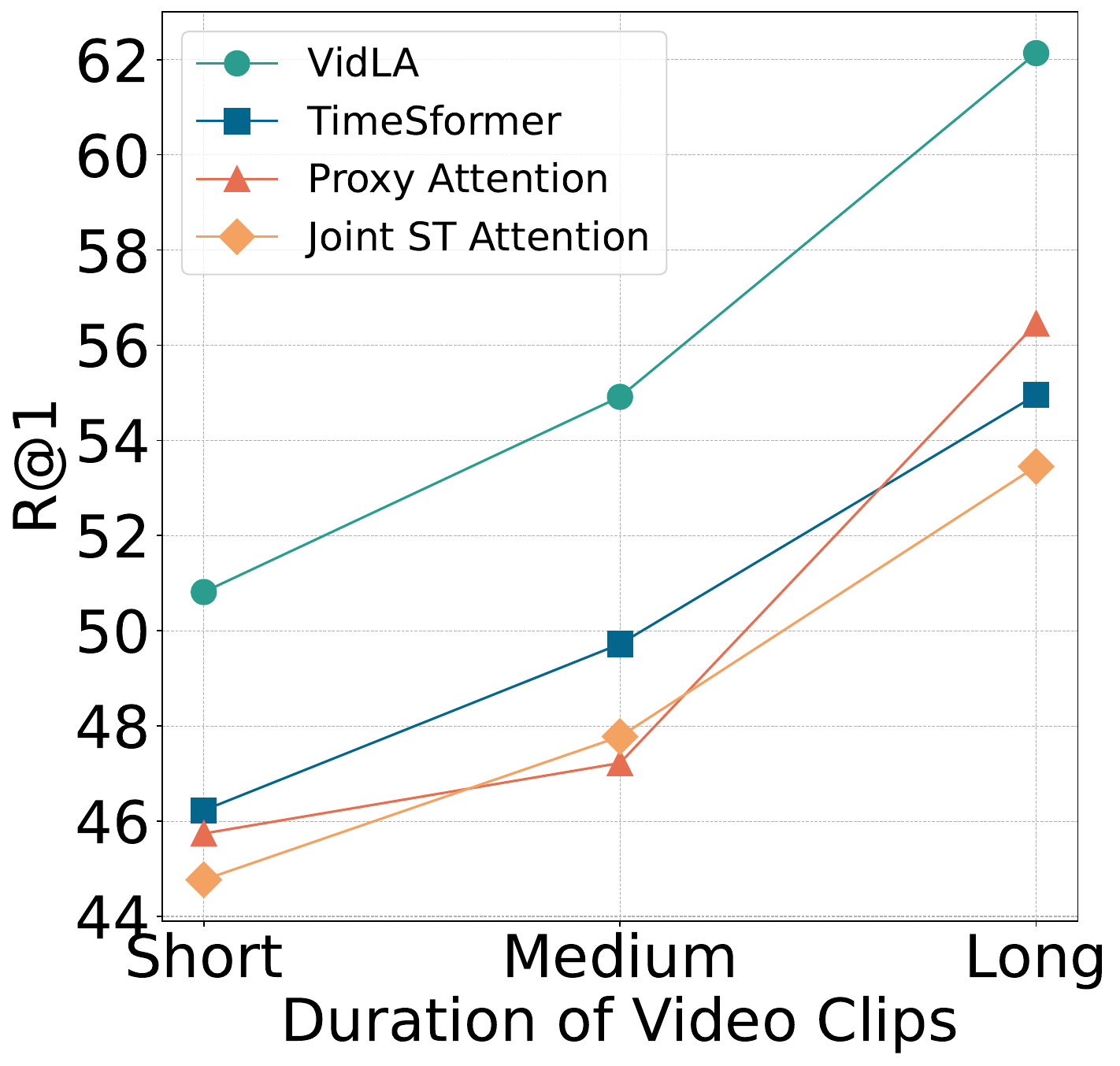}{\label{fig:attention_duration}} }}
    \subfloat[]{{\includegraphics[width=0.24\textwidth]{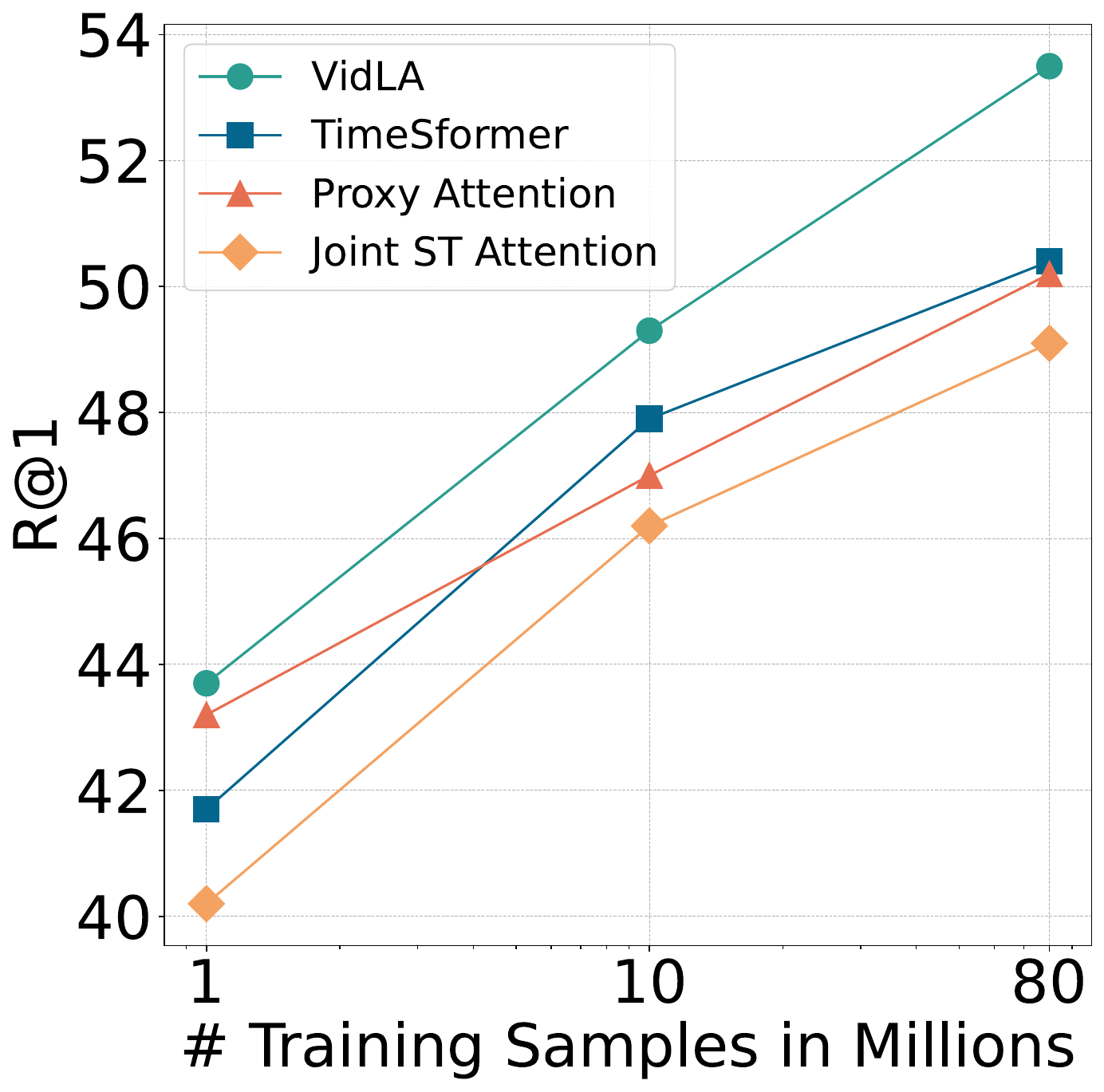}}{\label{fig:attention_scale}} }
    \vspace{-6mm}
    \caption{\small Retrieval performance on MSR-VTT compared to other attention mechanisms \textit{Left}: R@1 numbers for validation videos separated into 3 bins of different durations. VidLA consistently improves over baselines for all video durations. \textit{Right}: Scaling up the pretraining dataset improves the performance. Our architecture improves over other attention mechanisms at all data scales.
    }
    \label{fig:attention_analysis}%
\vspace{-5mm}
\end{figure}

\noindent\textbf{Retrieval Performance on Videos of Different Lengths}
To conduct a more finegrained analysis of the performance of our method, in the left plot of Figure~\ref{fig:attention_analysis}, we compare the performances of VidLA with respect to other attention methods on videos of different lengths. For this analysis, we report MSR-VTT R@1 results for three splits of videos in the validation set. Particulalry, we sort the videos by length and pick the shortest third for the short split, longest third for the long split and the remaining for the medium split. We observe that VidLA consistently outperforms other methods on all splits of different video lengths.

\begin{table}[h]
\begin{center}
\small
\begin{tabular}{ccc|cccc}
\hline

\hline

\hline\\[-3mm]
{\multirow{2}{*}{Sub}} &{\multirow{2}{*}{Cap}} & {\multirow{2}{*}{Sum}} & \multicolumn{4}{c}{MSR-VTT Retrieval}\\
& & & {R@1} & {R@5} & {R@10} & Avg\\
\hline
\cmark & \cmark & \xmark & 36.3 & 65.0 & 76.3 & 59.2 \\
\xmark & \cmark & \cmark & 48.9 & 74.1 & 84.0 & 69.0 \\
\cmark & \xmark & \cmark & 50.1 & 76.7 & 84.5 & 70.4 \\
\rowcolor{blue!5} \cmark & \cmark & \cmark & \textbf{53.5} & 77.5	& 85.6	& \textbf{72.2}  \\
\hline 

\hline

\hline
\end{tabular}
\end{center}
\vspace{-1.2em}
\caption{Comparison of finetuned retrieval performances on MSR-VTT dataset with different language supervision during pretraining. We compare the effectiveness of using subtitles, captions and whether or not they are summarized.}
\label{tab:language_sup}
\vspace{-0.5em}
\end{table}

\noindent\textbf{Training Data Size}
It is well-known that performance of retrieval models scales with the pretraining data size in the contrastive learning setting. We study our model's performance as a function of the pretraining dataset size by pretraining different models on datasets of sizes 80M, 10M and 1M. We report the results in the right plot on Figure~\ref{fig:attention_analysis} and compare the performance of VidLA with other attention methods. We notice that VidLA outperforms all the methods across all data scales.

\noindent\textbf{Effect of Different Language Supervision}
To validate the efficacy of utilizing both subtitles and captions for language supervision, as well as the effectiveness of text summarization, we pretrain our model with different combinations of text sources and summarization. From the results presented in Table~\ref{tab:language_sup}, we observe that the model's performance is better with supervision from both subtitles and captions compared to using only one of the two. Additionally, removing summarization significantly degrades performance. Without summarization, video-text alignment suffers due to increased verbosity in longer videos and the inability to leverage CLIP's pretrained embedding layer due to increased context length.

\begin{table}[ht!]
\begin{center}
\resizebox{0.48\textwidth}{!}{
\small
\begin{tabular}{lc|cc|cc}
\hline

\hline

\hline\\[-3mm]
 \multicolumn{1}{l}{\multirow{2}{*}{{Method}}} & {\multirow{2}{*}{{Frames}}}&
 \multicolumn{2}{c|}{{K400}} & \multicolumn{2}{c}{{Sth-sth-v2}}\\  
\multicolumn{1}{c}{} & & {Views} & {Top-1} & {Views} & {Top-1} \\
\hline
TimeSformer-B/16~\cite{bertasius2021space}  & 96 & $1 \times 3$ & 80.7 & $1 \times 3$ &62.4 \\
VideoMAE-B/16~\cite{tong2022videomae} & 16 & $5\times3$ & 81.5	&$2 \times 3$ & 70.8 \\
VideoMAE-v2-B/16~\cite{wang2023videomaev2} & 16 & $5\times3$ & 81.5	&$2 \times 3$ & \textbf{71.2} \\
ViViT-L/16~\cite{arnab2021vivit} & 32 & $1 \times 3$ & 81.7 & $1 \times 1$ &65.9 \\
VideoSwin-B~\cite{Liu_2022_CVPR} & 32 & $3 \times 4$ & 82.7 & $1 \times 3$ & 69.6 \\
UMT-B/16$_{800e}$~\cite{li2023unmasked} & 8 & $3 \times 4$ & \textbf{85.7} & $2 \times 3$ &70.8 \\
\rowcolor{blue!5} VidLA-B/32 & 16 & $5 \times 3$ & 82.4 &	$2 \times 3$	& 67.9 \\
\rowcolor{blue!5} VidLA-B/16 & 16 & $5 \times 3$ & 84.9 &	$2 \times 3$	& 69.9 \\
\hline 

\hline

\hline
\end{tabular}   
}
\end{center}
\vspace{-1.6em}
\caption{Comparison of finetuned classification performances on Kinetics-400 and Something-Something-v2. VidLA models using ViT-B backbones achieve competitive results in spite of being pretrained only for alignment.}
\label{tab:classification_k400_ssv2}
\vspace{-0.5em}
\end{table}

\noindent\textbf{Classification Results} 
Even though our proposed method primarily focuses on video-language alignment, we evaluate the performance of our method on a related downstream task, \ie, action recognition. We add a classification head on top of the video encoder from VidLA and finetune it on the popular benchmark datasets Kinetics-400~\cite{kay2017kinetics} and Something-Something-V2~\cite{goyal2017something}. We report the results of the finetuned models in Table~\ref{tab:classification_k400_ssv2}. Although VidLA was pretrained only for video-language alignment, we observe that VidLA performs competitively even against models such as VideoMAE that use dense pretraining objectives to promote the learning of finegrained features.

\section{Conclusion}
In this work, we propose a novel hierarchical temporal modeling architecture that captures temporal relationships at multiple temporal scales while remaining flexible to leverage image-text pretrained models. We also introduce an approach for utilizing LLMs to create the largest video-language dataset with better semantic alignment between video and language. We empirically validate the efficacy of our proposed hierarchical temporal attention mechanism as well as its design choices on data with varying temporal lengths and at different dataset sizes, demonstrating its advantage over other performant temporal modeling approaches. Our extensive experimentation also validates our data curation choices. Overall, our results highlight the importance of both high-quality large-scale training data as well as simple and scalable temporal architecture, and establishes VidLA as the new state-of-the-art on multiple video-text retrieval benchmarks while demonstrating its competitiveness on classification benchmarks.

{\small
\bibliographystyle{ieee_fullname}
\bibliography{egbib}
}

\clearpage
\setcounter{page}{1}
\appendix

\section{Implementation details}
\label{sec:implementation}
In this section, we provide additional implementation details of pretraining as well finetuning for both retrieval and classification tasks. We also include additional details regarding YT-VidLA-800M creation. 

\subsection{Additional pretraining details}
For pretraining, we set the weight decay to $0.02$. We use AdamW~\cite{loshchilov2017decoupled} optimizer and set $\beta_1$ and $\beta_2$ to $0.9$ and $0.999$ respectively. We set gradient clipping to $10$.  

\subsection{Finetuning details for retrieval}
For retrieval finetuning on all downstream datasets, we utilize 32 frames. For updating model parameters, we utilize AdamW optimizer with an initial learning rate of $1e-5$ and decay it to $1e-6$ using cosine annealing. We set the weight decay to $0.2$. For VidLA-B/32 finetuning on all datasets, we set the batch size to $128$, whereas, for VidLA-B/16 finetuning, we set the batchsize to 196. Unlike pretraining, we do not perform any multi-scale spatial crop for finetuning, rather, we perform center crop while preserving aspect ratio. We train VidLA-B/32 model for $440$ steps on MSVD dataset. We respectively train VidLA-B/32 for $2.5\times$, $2.5\times$, $2\times$, and $6\times$ steps on MSR-VTT, DiDeMo, ActivityNet Captions, and VATEX datasets. For VidLA-B/model, we train for $4\times$, $1.5\times$, $2\times$, $1\times$, and $6\times$ steps on MSR-VTT, DiDeMo, ActivityNet Captions, MSVD and VATEX datasets, respectively. 

\subsection{Finetuning details for classification}
For finetuning the video encoders from the pre-trained VidLA-B/32 for classification, we train for $100$ epochs with a  batch size of $512$ for Kinetics-400 and $40$ epochs with batch size $1024$ for Something-Something-v2. For VidLA-B/16, we train for $75$ epochs and $40$ epochs respectively for Kinetics-400 and Something-Something-v2, each with a  batch size of $512$. We use the dense sampling approach of \cite{feichtenhofer2019slowfast} for Kinetics-400, and TSN \cite{wang2019TSN} uniform sampling for Something-Something-v2.

\subsection{Additional details for YT-VidLA-800M creation}
For caption generation, we use Blip-2 ViT-L OPT\textsubscript{2.7B}~\cite{li2023blip}. We utilize Llama-2 13B Chat~\cite{touvron2023llama} model for caption and ASR subtitle summarization. We provide the summarization prompt in the following.

\medskip 

\texttt{Summarize the following sentences into a single sentence, not exceeding 25 words. Do not output any additional text and use any external information.}
\texttt{<input text>}

\medskip

\section{Zero-Shot Retrieval}
\label{sec:zs_retrieval}
We evaluate the zero-shot retrieval performance of VidLA-B/16 on the MSR-VTT dataset. We present the results in Table~\ref{tab:zs_retrieval}. We compare the performance with both two-tower and two-stage methods, without specific constraints on vision backbone size (e.g., ViT-B/16). We observe that VidLA outperforms all the prior works by a significant margin that employ similar-sized architectures. To be more specific, VidLA outperforms the recent two-stage method UMT~\cite{li2023unmasked} by 6\%. Surprisingly, we also find that VidLA outperforms InternVideo~\cite{wang2022internvideo} by 0.8\%, even though InternVideo utilizes a larger vision backbone. These results provide further evidence for the effectiveness of VidLA in the video-language alignment task.

\begin{table}[ht]
\begin{center}
\small
\begin{tabular}{lc}
\hline

\hline

\hline\\[-3mm]
Method & R@1\\
\hline
Frozen~\cite{bain2021frozen} & 18.7 \\
VIOLET~\cite{fu2021violet} & 25.9 \\
CLIP4Clip~\cite{luo2022clip4clip} & 30.6 \\ 
VINDLU{$\dagger$}~\cite{cheng2023vindlu} & 32.0 \\
Singularity~\cite{lei2022revealing} & 34.0  \\
OmniVL~\cite{wang2022omnivl} & 34.6 \\ 
VideoCoCa~\cite{yan2022videococa} & 34.3 \\
UMT-B{$\dagger$}~\cite{li2023unmasked} & 35.5 \\
InternVideo(ViT-L)*~\cite{wang2022internvideo} & 40.7 \\
\rowcolor{blue!5} VidLA-B/16 & 36.6  \\
\rowcolor{blue!5} VidLA-B/16* & \textbf{41.5}  \\
\hline 

\hline

\hline
\end{tabular}
\end{center}
\vspace{-1.2em}
\caption{Zero-shot retrieval results on MSR-VTT. * indicates inference with dual-softmax. {$\dagger$} indicates two-stage method with candidate re-ranking.}
\label{tab:zs_retrieval}
\vspace{-4mm}
\end{table}

\section{Zero-Shot Multiple Choice}
\label{sec:zs_multichoice}

The objective of the zero-shot multiple choice task is to find the correct answer from a list of options. We evaluate the performance of VidLA-B/16 on the MSR-VTT dataset for this task and report the results in Table~\ref{tab:zs_multichoice}. We observe that VidLA outperforms the previous state-of-the-art, InternVideo, by 0.9\%. We also notice that VidLA outperforms All-in-one by 11.9\%. Overall, these results further demonstrate the generalizability of our large-scale video-language alignment training.

\begin{table}[ht]
\begin{center}
\small
\begin{tabular}{lc}
\hline

\hline

\hline\\[-3mm]
Method & Accuracy\\

\hline
All-in-one~\cite{wang2023all} & 80.3 \\
InternVideo(ViT-B)~\cite{wang2022internvideo} & 91.3 \\
\rowcolor{blue!5} VidLA-B/16 & \textbf{92.2}  \\
\hline 

\hline

\hline
\end{tabular}
\end{center}
\vspace{-1.2em}
\caption{Zero-shot multiple choice results on MSR-VTT.}
\label{tab:zs_multichoice}
\vspace{-4mm}
\end{table}

\section{Zero-Shot Classification}
\label{sec:zs_classification}
We evaluate the performance of VidLA-B/16 on the zero-shot action classification task using the Kinetics-400 dataset. We only compare with methods that do not utilize Kinetics-400 data (labels and videos) for pretraining. To compare with prior art, we report results on both validation and test sets of Kinetics-400. We perform a `single-view' evaluation and utilize uniformly sampled 32 frames. For evaluation, following CLIP~\cite{radford2021learning}, we obtain the embedding of all the class labels using the text encoder while utilizing the default prompt templates for Kinetics-400 classification. 
We consider the prediction as correct if the groundtruth label has the maximum similarity with the video embedding. We report the results in Table~\ref{tab:zs_classification}. We observe that our proposed method outperforms all prior works by a noticeable margin. To be precise, VidLA outperforms CLIP by around 11\% on both validation and test sets. VidLA also outperforms, ActionCLIP~\cite{wang2021actionclip} by 3.8\%. We also notice that VidLA outperforms current state-of-the-art VFC~\cite{momeni2023verbs} on both validation and test sets. Note that VFC~\cite{momeni2023verbs} employs a verb focused contrastive loss during pretraining which greatly improves performance on action recognition task since the extracted verbs are, in principle, very similar to the downstream action recognition labels. Therefore, we also report a baseline score from VFC~\cite{momeni2023verbs} which does not employ such a verb focused contrastive objective. VidLA outperforms the VFC-baseline by 3.5\% on the Kinetics-400 test set. These results further validate the effectiveness of VidLA in action recognition, complementing the finetuned results reported in Table 7 of the main text.

\begin{table}[ht]
\begin{center}
\small
\begin{tabular}{lc}
\hline

\hline

\hline\\[-3mm]
Method & Top-1 Accuracy\\

\hline
\rowcolor{gray!5} \multicolumn{2}{c}{Validation-Set} \\
CLIP~\cite{radford2021learning} & 48.9 \\
ActionCLIP~\cite{wang2021actionclip} & 56.4 \\
VFC~\cite{momeni2023verbs} & 59.4 \\
\rowcolor{blue!5} VidLA-B/16	& \textbf{60.2}  \\
\hline
\rowcolor{gray!5} \multicolumn{2}{c}{Test-Set} \\
Flamingo-3B~\cite{alayrac2022flamingo} & 45.2 \\
Flamingo-80B~\cite{alayrac2022flamingo} & 49.1 \\
Flamingo-9B~\cite{alayrac2022flamingo} & 49.7 \\
CLIP~\cite{radford2021learning} & 47.9 \\
VFC-Baseline~\cite{momeni2023verbs} & 55.6 \\
VFC~\cite{momeni2023verbs} & 58.8 \\
\rowcolor{blue!5} VidLA-B/16	& \textbf{59.1}  \\
\hline 

\hline

\hline
\end{tabular}
\end{center}
\vspace{-1.2em}
\caption{Zero-shot classification results on Kinetics-400.}
\label{tab:zs_classification}
\vspace{-4mm}
\end{table}

\section{Video-to-Text Retrieval Results}
\label{sec:v2t_result}
We present video-to-text retrieval results on DiDeMo, ActivityNet Captions, MSVD, and VATEX datasets in Table~\ref{tab:v2t_didemo}, \ref{tab:v2t_actnet}, \ref{tab:v2t_msvd}, and \ref{tab:v2t_vatex} respectively. Similar to the text-to-video retrieval results reported in the main text, we observe that VidLA outperforms prior art on video-to-text retrieval by a significant margin in most settings. However, InternVideo outperforms VidLA-B/16 on the VATEX dataset, likely due to its larger backbone architecture.

\begin{table}[h]
\begin{center}
\small
\begin{tabular}{l|cccc}
\hline

\hline

\hline\\[-3mm]
{\multirow{2}{*}{Method}} & \multicolumn{4}{c}{DiDeMo Retrieval}\\
 & {R@1} & {R@5} & {R@10} & Avg \\

\hline
\rowcolor{gray!5} \multicolumn{5}{c}{CLIP-ViT-B/32} \\
CLIP4Clip~\cite{luo2022clip4clip}   & 42.5 & 70.6 & 80.2 & 64.4 \\
CAMoE*~\cite{cheng2021improving}    & 45.5 & 71.2 & $-$ & $-$  \\
DRL~\cite{wang2022disentangled} & 45.4 & 72.6 & 82.1 & 66.7  \\
DiffusionRet*~\cite{jin2023diffusionret} & 50.3 & 75.1 & 82.9 & 69.4 \\
\rowcolor{blue!5} VidLA-B/32 & 55.0 & 82.8	& 88.9 & 75.6  \\
\rowcolor{blue!5} VidLA-B/32* & \textbf{62.2} & 85.1 & 91.1 & \textbf{79.5}  \\
\rowcolor{gray!5} \multicolumn{5}{c}{CLIP-ViT-B/16} \\
DRL~\cite{wang2022disentangled} & 49.9 & 75.4 & 83.3 & 69.5\\
InternVideo(ViT-L)* & 59.1 & $-$ & $-$ & $-$ \\
UMT{$\dagger$} & 59.5 & 84.9 & 90.5 & 78.3  \\
\rowcolor{blue!5} VidLA-B/16 & 59.6 & 84.4	& 90.2 & 78.0  \\
\rowcolor{blue!5} VidLA-B/16* & \textbf{67.5} & 87.5 & 91.7 & \textbf{82.2}  \\

\hline 

\hline

\hline
\end{tabular}
\end{center}
\vspace{-1.2em}
\caption{Video-to-text retrieval on DiDeMo dataset. * indicates inference with dual-softmax. {$\dagger$} indicates two-stage method with candidate re-ranking.}
\label{tab:v2t_didemo}
\vspace{-4mm}
\end{table}

\begin{table}[h]
\begin{center}
\small
\begin{tabular}{l|cccccc}
\hline

\hline

\hline\\[-3mm]
{\multirow{2}{*}{Method}} & \multicolumn{4}{c}{ActivityNet Captions Retrieval}\\
 & {R@1} & {R@5} & {R@10} & Avg\\

\hline
\rowcolor{gray!5} \multicolumn{5}{c}{CLIP-ViT-B/32} \\
CLIP4Clip~\cite{luo2022clip4clip}   & 42.5 & 74.1 & 85.8 & 67.5 \\
CenterCLIP~\cite{zhao2022centerclip}  & 44.5 & 75.7 & 86.2 & 68.8 \\
DRL~\cite{wang2022disentangled} & 42.2 & 74.0 & 86.2 & 67.5 \\
CAMoE*~\cite{cheng2021improving}    & 49.9 & 77.4 & $-$ & $-$ \\
DiffusionRet*~\cite{jin2023diffusionret} & 47.4 & 76.3 & 86.7 & 70.1\\

\rowcolor{blue!5} VidLA-B/32 & 60.8 & 85.0	& 91.8 & 79.2  \\
\rowcolor{blue!5} VidLA-B/32* & \textbf{69.1} & 88.8 & 93.7 & \textbf{83.9}  \\
\rowcolor{gray!5} \multicolumn{5}{c}{CLIP-ViT-B/16} \\
DRL~\cite{wang2022disentangled} & 45.7 & 76.5 & 87.8 & 70.0 \\
CenterCLIP~\cite{zhao2022centerclip}  & 46.7 & 77.1 & 88.0 & 70.6 \\
UMT{$\dagger$} & 56.0 & 83.5 & 91.7 & 77.1  \\
InternVideo(ViT-L)* & 62.8 & $-$ & $-$ & $-$  \\
\rowcolor{blue!5} VidLA-B/16 & 63.4 & 87.5 & 93.2 & 81.4 \\
\rowcolor{blue!5} VidLA-B/16* & \textbf{72.9} & 90.6 & 94.9 & \textbf{86.1} \\

\hline 

\hline

\hline
\end{tabular}
\end{center}
\vspace{-1.2em}
\caption{Video-to-text retrieval on ActivityNet Captions dataset. * indicates inference with dual-softmax. {$\dagger$} indicates two-stage method with candidate re-ranking.}
\label{tab:v2t_actnet}
\vspace{-4mm}
\end{table}

\begin{table}[h]
\begin{center}
\small
\begin{tabular}{l|cccc}
\hline

\hline

\hline\\[-3mm]
{\multirow{2}{*}{Method}} & \multicolumn{4}{c}{MSVD Retrieval}\\
 & {R@1} & {R@5} & {R@10} & Avg \\

\hline

\rowcolor{gray!5} \multicolumn{5}{c}{CLIP-ViT-B/32} \\
CLIP4Clip~\cite{luo2022clip4clip}   & 62.0 & 87.3 & 92.6 & 80.6 \\

DRL~\cite{wang2022disentangled} & 62.3 & 86.3 & 92.2 & 80.3  \\
CAMoE*~\cite{cheng2021improving}    & 49.8 & 79.2 & 87.0 & 72.0  \\
DiffusionRet~\cite{jin2023diffusionret} & 61.9 & 88.3 & 92.9 & 81.0 \\
CenterCLIP~\cite{zhao2022centerclip}  & 63.5 & 86.4 & 92.6 & 80.8 \\

\rowcolor{blue!5} VidLA-B/32 & 66.5 & 92.4	& 95.6 & 84.8 \\
\rowcolor{blue!5} VidLA-B/32* & \textbf{77.4} & 96.0 & 98.7 & \textbf{90.7}  \\
\rowcolor{gray!5} \multicolumn{5}{c}{CLIP-ViT-B/16} \\
CenterCLIP~\cite{zhao2022centerclip}  & 68.4 & 90.1 & 95.0 & 84.5 \\
DRL~\cite{wang2022disentangled} & 68.7 & 92.5 & 95.6 & 85.6 \\
InternVideo(ViT-L)* & 76.3 & $-$ & $-$ & $-$ \\
UMT{$\dagger$} & 74.0 & 94.6 & 97.3 & 88.6 \\
\rowcolor{blue!5} VidLA-B/16 & 68.8 & 93.4 & 96.2 & 86.1  \\
\rowcolor{blue!5} VidLA-B/16* & \textbf{80.3} & 97.2 & 98.4 & \textbf{92.0}\\

\hline 

\hline

\hline
\end{tabular}
\end{center}
\vspace{-1.2em}
\caption{Video-to-text retrieval on MSVD dataset. * indicates inference with dual-softmax. {$\dagger$} indicates two-stage method with candidate re-ranking.}
\label{tab:v2t_msvd}
\vspace{-4mm}
\end{table}

\begin{table}
\begin{center}
\small
\begin{tabular}{l|cccc}
\hline

\hline

\hline\\[-3mm]
{\multirow{2}{*}{Method}} & \multicolumn{4}{c}{VATEX Retrieval}\\
 & {R@1} & {R@5} & {R@10} & Avg\\
\hline

\rowcolor{gray!5} \multicolumn{5}{c}{CLIP-ViT-B/32} \\

DRL~\cite{wang2022disentangled} & 77.0 & 98.0 & 99.4 & 91.5 \\
\rowcolor{blue!5} VidLA-B/32 & 79.8 & 99.5 & 99.8 & 93.0 \\
\rowcolor{blue!5} VidLA-B/32* & \textbf{85.0} & 99.8 & 99.9 & \textbf{94.9} \\
\hline
\rowcolor{gray!5} \multicolumn{5}{c}{CLIP-ViT-B/16} \\
DRL~\cite{wang2022disentangled} & 80.1 & 98.5 & 99.5 & 92.7 \\
InternVideo(ViT-L)* & \textbf{87.2} & $-$ & $-$ & $-$ \\
\rowcolor{blue!5} VidLA-B/16 & 81.1 & 99.5 & 99.9 & 93.5  \\
\rowcolor{blue!5} VidLA-B/16* & 85.6 & 99.8 & 100.0 & \textbf{95.1} \\

\hline 

\hline

\hline
\end{tabular}
\end{center}
\vspace{-1.2em}
\caption{Video-to-text retrieval on VATEX dataset. * indicates inference with dual-softmax.}
\label{tab:v2t_vatex}
\vspace{-4mm}
\end{table}

\section{Effect of Image-Language Pretraining}
\label{sec:pretraining}
\begin{figure}[ht!]
\begin{center}
\includegraphics[width=0.9\columnwidth]{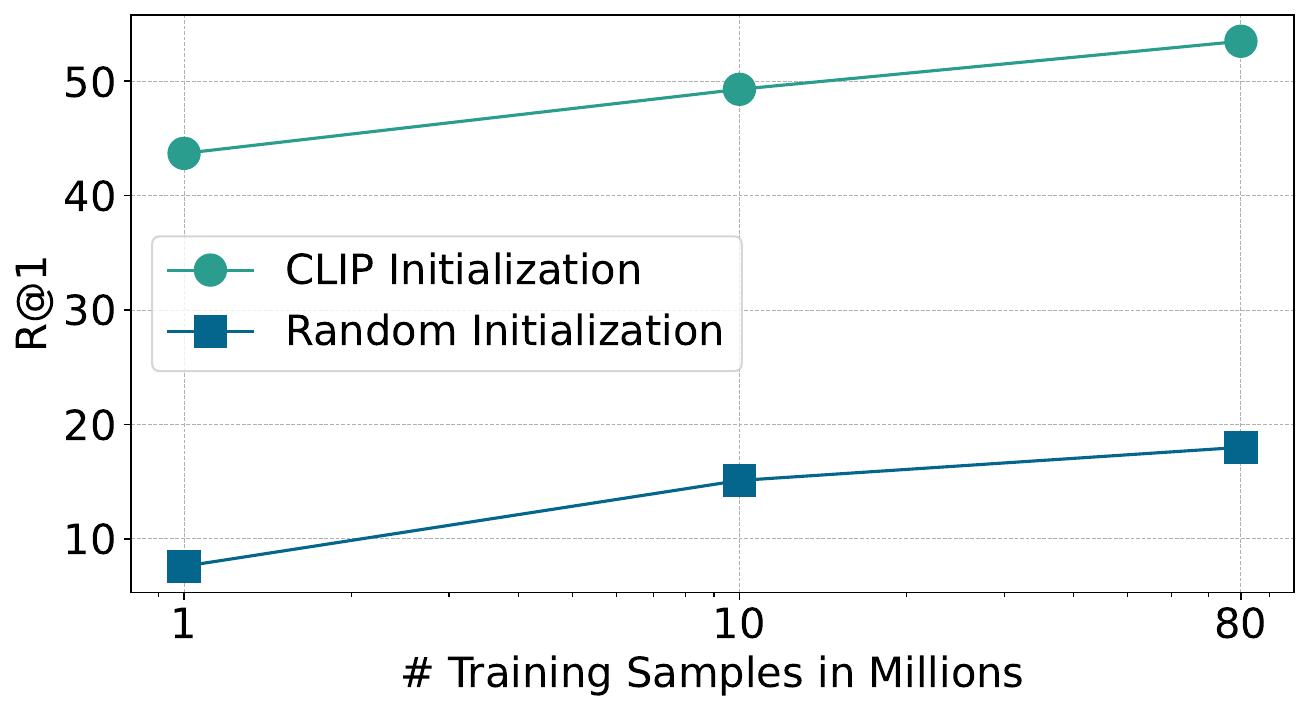}
\caption{Effect of image-language pretraining on MSR-VTT.}
\label{fig:pretrained_vs_scratch}
\end{center}
\vspace{-4mm}
\end{figure}

A key objective of our work is to introduce minimal architectural modifications to effectively utilize pretrained image-language models. To analyze the effectiveness of image-language pretraining, we conduct experiments where we initialize the network weights randomly instead of using the CLIP weights. We evaluate the performance on three data scales and present the retrieval results on the MSR-VTT dataset in Figure~\ref{fig:pretrained_vs_scratch}. Figure~\ref{fig:pretrained_vs_scratch} demonstrates that initializing the model weights with CLIP weights significantly outperforms the case where we initialize the model weights randomly at all data scales. This finding further validates the efficacy of VidLA in effectively utilizing the pretrained image-language weights and highlights the challenge of training from scratch for video-language alignment, which requires a large amount of training data.         

\section{Analysis of \concept Token Configuration}
\label{sec:mst_design}
To investigate the impact of various design aspects of \concept tokens, we systematically vary the number of hierarchies, $U$, and the temporal scale, $r$ and evaluate the performance on the MSR-VTT dataset. We conduct these experiments on the 80 million data scale, consistent with most of the analysis done in the main text. We report the results in Table~\ref{tab:mst_design}.

The first and the second rows of Table~\ref{tab:mst_design} demonstrate that incorporating \concept tokens for a single hierarchy, $U=1$, leads to noticeable performance gains. The third and fourth rows further illustrate that increasing the number of hierarchies to 2 provides additional improvements, and setting the temporal scale to a higher value leads to better results. Finally, the last two rows indicate that adding one more hierarchy provides further improvement. However, we do not notice a significant difference between different temporal scales. Note that the \concept token design in the last row is our default setting. The results in Table~\ref{tab:mst_design} provides further support for our hypothesis that incorporating hierarchy into the design of \concept tokens plays a crucial role in improving video-language alignment performance.    

\begin{table}[h]
\begin{center}
\small
\begin{tabular}{c|cccc}
\hline

\hline

\hline\\[-3mm]
{\multirow{2}{*}{$U$-$V$-$r$}} & \multicolumn{4}{c}{MSR-VTT Retrieval}\\
 & {R@1} & {R@5} & {R@10} & Avg\\

\hline
\xmark & 49.1 & 75.3 & 83.5 & 69.3 \\
1-4-1 & 51.3 & 76.5 & 85.0 & 70.9 \\
2-4-2 & 52.1 & 76.6 &	84.2 &	71.0 \\
2-4-3 & 52.3	& 76.9 & 85.7 & 71.6 \\
3-4-3 & 53.2	& 77.0 & 85.6 &	71.9 \\
\rowcolor{blue!5} 3-4-2  & \textbf{53.5} & 77.5	& 85.6	& \textbf{72.2}  \\
\hline 

\hline

\hline
\end{tabular}
\end{center}
\vspace{-1.2em}
\caption{Retrieval performance on MSR-VTT with different \concept designs.}
\label{tab:mst_design}
\vspace{-4mm}
\end{table}

\section{Results on Hierarchical Dataset}
We demonstrated VidLA's efficacy on both short and long videos in Figure 4 (left). To further illustrate VidLA's effectiveness in a strictly hierarchical setting, we constructed a temporally hierarchical test set with 1K short, mid and long clip triplets centered around the same timestamp. Zero-shot retrieval performance in Table~\ref{tab:hierarchical_ret} shows VidLA outperforms current state-of-the-art method, CLIPViP, even when trained on the same pretraining dataset (YT-VidLA-80M).

\begin{table}[h]
\centering
    \begin{tabular}{lccc}
    \hline
Method & Short & Medium & Long\\
\hline
CLIPViP-B/32 & 85.9 & 86.8 & 88.1 \\
\rowcolor{blue!5} VidLA-B/32 & 86.8 & 90.5 & 92.5 \\
\hline
    \end{tabular}
    \vspace{-0.7em}
    \vspace{-2pt}
    \caption{Text2Video retrieval R@1 on a 1k hierarchical test sets}
    \label{tab:hierarchical_ret}
\end{table}

\section{Analysis of Pretraining Dataset}

To analyze the effect of different pretraining datasets, we pretrain VidLA-B/32 on diffrent pretraining datasets and later finetune on downstream datasets and report the results in Table~\ref{tab:dataset_analysis}. The first row reports the results where no pretaining on video data was performed and the model was only finetuned on the downstream dataset. From second and third row we notice that training on YT-VidLA-10M outperforms training on WebVid-10M. The last two rows demonstrate the advantage of our large-scale training where the performance improves significantly with scale.

\begin{table}[h]
\begin{center}
\small
\begin{tabular}{lccc}
    \hline
Dataset & MSR-VTT & DiDeMo & ActivityNet\\
\hline
WIT-400M & 42.9 & 40.7 & 40.7 \\
WebVid-10M & 43.6 & 46.6 & 46.9 \\
\rowcolor{blue!5} YT-VidLA-10M & 49.3 & 46.3 & 49.7 \\
\rowcolor{blue!5} YT-VidLA-80M & 53.5 & 52.3 & 53.9  \\
\rowcolor{blue!5} YT-VidLA-800M & 55.6 & 56.9 & 61.3  \\
\hline
\end{tabular}
\end{center}
\vspace{-1.2em}
\caption{Text-to-Video retrieval R@1 with different pretraining datasets}
\label{tab:dataset_analysis}
\end{table}

\end{document}